\documentclass[letterpaper, 10pt, conference, dvipsnames]{ieeeconf}
\IEEEoverridecommandlockouts

\usepackage[pdftex]{graphicx}
\usepackage{multicol}
\usepackage{amsmath, amsfonts, amssymb}
\usepackage{xspace}
\usepackage{latexsym}
\usepackage{url}
\usepackage{listings}
\usepackage{float}
\usepackage{tabularx}
\usepackage{tabulary}
\usepackage{tikz}
\usepackage{pgfplots}
\usepackage{fancybox}
\usepackage{flushend}
\usepackage{lipsum}
\usepackage[ruled,lined]{algorithm2e}
\usepackage{hyphenat} 
\usepackage{xcolor}

\lstdefinestyle{Prolog} {language=Prolog,
                         lineskip=-0.3ex,
                         fontadjust=true,
                         basicstyle={\footnotesize\nopagebreak[4]},
                         commentstyle=\footnotesize, 
                         keywordstyle=\footnotesize,
                         showstringspaces=false,
                         showspaces=false,
                         showtabs=false,
                         moredelim=**[is][\bf]{@}{@},
                         moredelim=**[is][\it]{~}{~}
                         }
\lstdefinestyle{Lisp}   {language=Lisp,
                         lineskip=-0.5ex,
                         fontadjust=true,
                         basicstyle={\footnotesize \nopagebreak[4]},
                         commentstyle=\footnotesize,
                         keywordstyle=\footnotesize,
                         moredelim=**[is][\it\bf\color{BrickRed}]{<}{<},
                         moredelim=**[is][\it\bf\color{Violet}]{&}{&},
                         moredelim=**[is][\it\bf\color{OliveGreen}]{^}{^},
                         moredelim=**[is][\it\bf\color{BurntOrange}]{!}{!},
                         moredelim=**[is][\bf]{@}{@},
                         moredelim=**[is][\it]{~}{~}
                         }

\usetikzlibrary{matrix}
\usepgfplotslibrary{groupplots}
\pgfplotsset{compat=newest}

\graphicspath{{../../../img/}}

\title{Specializing Underdetermined Action Descriptions Through Plan Projection}

\author{Gayane Kazhoyan and Michael Beetz*\\
        {\{kazhoyan, beetz\}@cs.uni-bremen.de}
\thanks{*The authors are with the Institute for Artificial Intelligence, 
University of Bremen, Germany.}}

\begin{document}
\maketitle

\begin{abstract}

Plan execution on real robots in realistic environments is underdetermined and often leads to failures.
The choice of action parameterization is crucial for task success. 
By thinking ahead of time with the fast plan projection mechanism proposed in this paper, a general plan can be specialized towards the environment and task at hand by
choosing action parameterizations that are predicted to lead to successful execution.
For finding causal relationships between action parameterizations and task success, we provide the robot with means for plan introspection
and propose a systematic and hierarchical plan structure to support that.
We evaluate our approach by showing how a PR2 robot, when equipped with the proposed system, is able to choose action parameterizations that increase task execution success rates and overall performance of fetch and deliver actions in a real world setting.

\end{abstract}


\section{Introduction}
\label{sec:intro}

There have been remarkable demonstrations of autonomous mobile manipulation robots performing everyday activities
such as folding clothes \cite{folding} or washing dishes \cite{dishes}.
However, the robot control programs for executing these tasks are only applicable in the specific
settings that they are implemented for.
To enable the robots escape laboratory settings and enter unstructured environments such as human households,
they need to be able to autonomously manipulate a big variety of objects 
in a multitude of task contexts, while dealing with differences in the environments
and constant failures due to inaccuracies in sensors, actuators and the world representation.

To generalize robot control programs towards different objects, tasks, environments and robot platforms,
we introduced the concept of \textit{entity descriptions} \cite{kazhoyan2017designators}.
These are abstract underspecified symbolic descriptions of task-relevant entities (objects, locations, actions, etc.)
that are being grounded during execution into robot's environment through perception and reasoning.
During grounding they are augmented with symbolic and subsymbolic data that specializes them to the environment at hand.
Here is an example action description:

\begin{lstlisting}[style=Lisp]
(an action (type picking-up)
           (object (the object (type cup) 
                               (pose ~a-pose~))))
\end{lstlisting}

\noindent which gets augmented with information such as which arm to pick up with and
with which trajectory:

\begin{lstlisting}[style=Lisp, caption={Grounded description of a pick up action}, captionpos=b, label={lst:desig}]
(an action (type picking-up)
           (object (the object (type cup)
                               (pose ~a-pose~)))
           @(arm left)@
           @(trajectory ~pose-1~ ~pose-2~ ...)@
           @...@)
\end{lstlisting}

\begin{figure}[thb]
\vspace{-3.5ex}
    \centering  
    \includegraphics[width=0.46\columnwidth]%
    {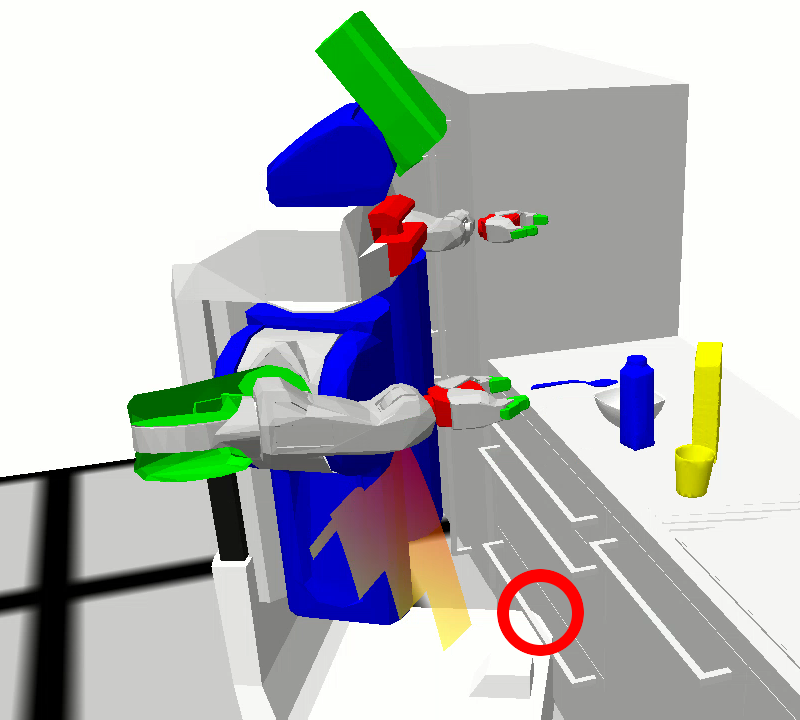}%
    \includegraphics[width=0.46\columnwidth]%
    {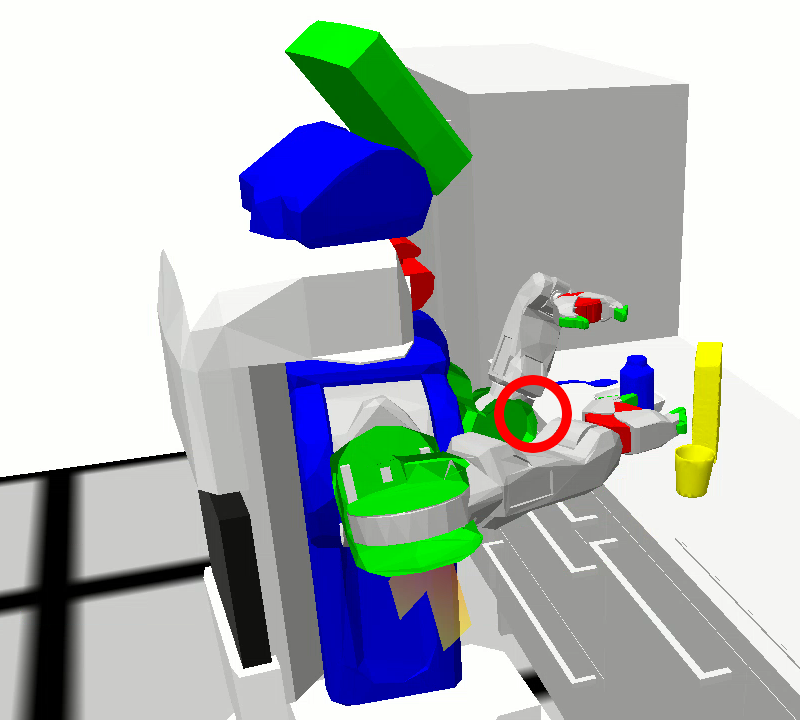}
    
    \hspace{0.25em}
    \includegraphics[width=0.46\columnwidth]%
    {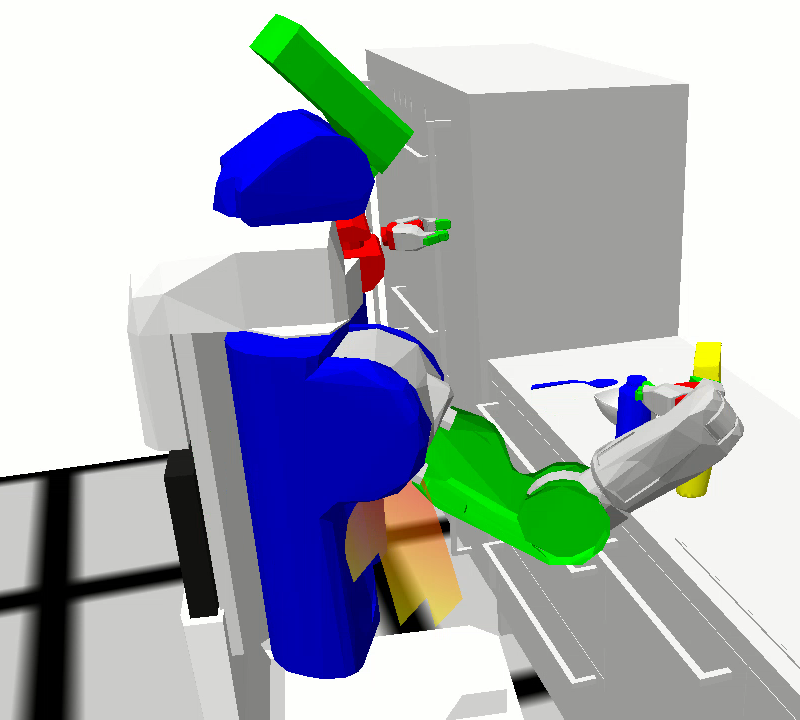}%
    \includegraphics[width=0.46\columnwidth]%
    {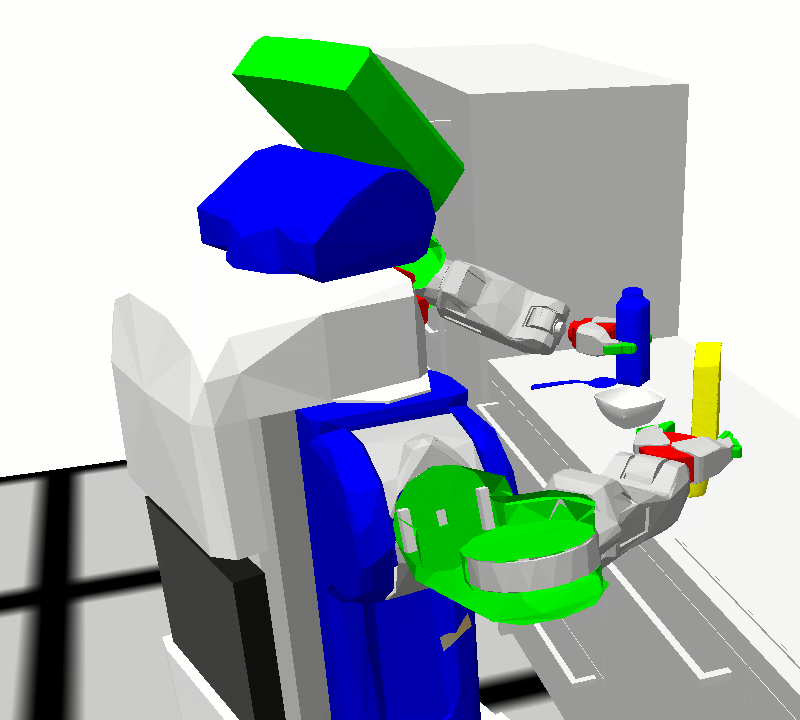} \newline
    \includegraphics[width=0.92\columnwidth]%
    {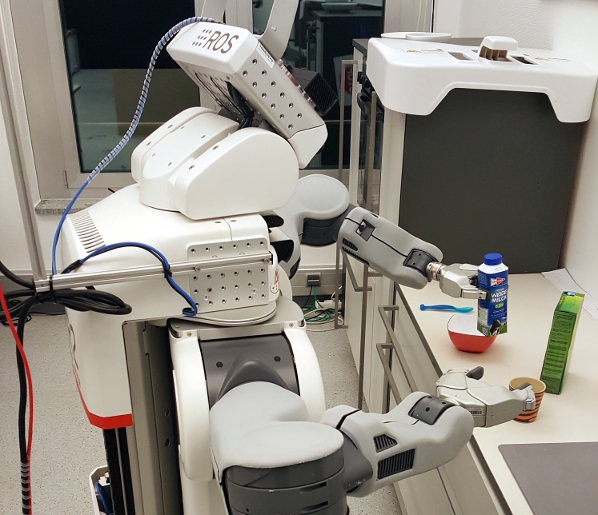}
  
  \caption{Projecting a plan for \textit{"fetch object of type milk"} action: \newline
  (top-left) step 1: robot approaches object -- base collides with furniture, \newline
  (top-right) step 2: backtracking, relocate and try left arm -- arm in collision, \newline
  (middle-left) step 3: try right arm -- no IK solution for next pose in trajectory, \newline
  (middle-right) step 4: relocate and try left arm again -- successful, \newline
  (bottom) step 5: execute chosen parameterization on real robot.}
  \label{fig:intro}
\end{figure}

There can be different groundings for the same entity description which result in different outcomes, including a variety of failures that can happen.
For example, for 
picking up a certain object, the chosen grasp type, robot base location, arm to use, the grasping force etc. decide if the action will be successful or, on the contrary, if the object will slip out, be out of reach, or if the trajectory will result in the robot colliding with the environment or knocking objects over (see Figure \ref{fig:intro}).

Robot control programs written with entity descriptions profit from a clean separation between the control flow and decision making.
The control flow specifies the ordering of actions and failure handling behaviors.
Decision making performs the reasoning necessary to find groundings and parameterizations of entities, which lead to successful task execution.
Introducing entity descriptions into programs makes them more scalable and generalizable towards different contexts
as well as more compact and readable, thereby decreasing code development and maintenance efforts.
However, the effort of specializing the program to the specific execution environment is thereby shifted from the programmer onto the robot itself.

This paper concentrates on reasoning required to adapt a general plan to a specific one tailored towards a known environment and scenario at hand
 by the means of reasoning into the future.
The adapting of the plan is implemented by specializing the underspecified entity descriptions contained in the plan. 
Intelligence here lies not in finding the correct sequence of actions but in the choice of action parameterizations that avoid failures and lead to successful 
task execution.
%
For example, the robot might need to answer a question such as: "If I grasp the cup from the handle with my left arm, will I be able to place it at the designated location without having to regrasp?" and based on the answers choose the arm to grasp with.
We present a fast plan projection mechanism\footnote{Open-source code and tutorials: \url{http://cram-system.org}}, which can be used to evaluate how well a plan with a particular parameterization will perform.
By thinking ahead of time through light-weight plan projection, the parameters of actions can be optimized by executing multiple runs of the same part of the plan with different parameterizations in the projection environment and choosing the best ones according to a cost function. The cost function can be based on the number of occurred handled or unhandled failures, on lengths of trajectories etc. Projection results can then be easily integrated into the robot's plan to execute the optimized part of the plan in the real world right away.

To enable the robot to infer if any of its executed actions were successful or triggered failures, i.e.\ to be able to find causal relationships between action parameterizations and task success, the robot needs to have means for introspection.
The importance of being able to reason about own execution has been argued, among others, by Ronald Brachman in his seminal talk titled "Systems That Know What They Are Doing" \cite{brachman}, where he explained how the ability of a system to comprehend what it is doing and why makes it robust in the face of unanticipated circumstances.
To implement introspection, we store plan-relevant information during real-world and projected execution into a data structure and provide the robot or its programmer with an interface for querying and reasoning with this knowledge.


Plans have to be well-structured such that relevant information --- action outcomes, failures that happened, parameterizations used, order of action execution etc.\ --- is available and easily retrievable to support introspection.
We propose systematic and hierarchical plan structure, based on the \textit{entity descriptions} concept \cite{kazhoyan2017designators}, and demonstrate how such plan structure is advantageous for introspection on the example of fetch and deliver plans.

The novel contributions of this paper are: 

\begin{itemize}
\item an approach to structure robot plans such that they are easily introspected;
\item a model for fetch and deliver plans and their implementation;
\item fast plan projection mechanism for a mobile manipulation robot as means to generate different behavior for the same symbolically-represented action;
\item mechanisms to incorporate plan projection online into a specific segment of the robot control program.
\end{itemize}

We evaluate our approach by showing how a PR2 robot is able to choose action parameterizations that increase task execution success rates and overall performance of fetch and deliver actions in a real world setting by using our system.


\section{Related Work}
\label{sec:related}

In cognitive science, predicting how motions affect the future evolution of the environment was found to play an essential role in human manipulation capabilities. An example of this line of thought is Hesslow's simulation theory of cognition \cite{hesslow2002conscious},
which proposes that thinking is imagined
interaction with the environment, during which behavior is simulated by activating motor and perceptual stimuli of the human but execution is suppressed such that it is not visible on the outside.
Szpunar et al.\ \cite{szpunar2014taxonomy} introduce the term \textit{prospection} to describe future-oriented cognition and propose a taxonomy thereof, which defines four modes of future thinking: simulation, prediction, intention and planning. These can include subsymbolic as well as semantic reasoning.
A brain-inspired cognitive architecture based on these ideas has been proposed by Shanahan \cite{cognitive-projection}.
It is based on a dynamical approach \cite{dynamical-cognitive} to cognitive science: in contrast to traditional approaches, which use language and symbolic reasoning as conceptual foundations for their cognitive architectures, the author makes use of an ``analog'' representation, which is realized through a neural network and is structurally closer to the perception-action domain of a cognitive agent. 
The agent runs parallel simulations with each action it can execute to estimate which one would bring to a state with highest reward.
The architecture is implemented with a large-scale neural network and is applied to a simple robot that can turn and perceive colorful cylinders.

In classical AI, symbolic plan projection is applied to predict the future state of the world. Projection is considered on the basis of axiomatized models of actions, which are atomic entities that have preconditions and effects. State-space planners, such as STRIPS \cite{ghallab2004automated} and more recent HTN-based planners such as SHOP2 \cite{shop2}, search through state transition systems with atomic transitions to find a sequence of actions, which is predicted to lead to the goal state. In the domain of mobile manipulation, the choice of action parameterization is crucial for task success. There is a large number of parameters and most of them are subsymbolic. Unfortunately, classical AI planners have a difficulty in handling such complex domains, therefore, they abstract away from motions. For example, an atomic grasping action is assumed to have an effect of an object necessarily being in hand after its execution, if certain preconditions have been met. In real world, grasping trajectories, reachability, occlusions, friction forces etc., which are abstracted away from in classical AI, are crucial for successful action execution.

On the other hand, also in classical AI prediction of the future has been considered as a useful tool and has been researched as a separate component of the planning system.
One of the first works in that direction is by Hanks \cite{hanks1990practical}, where he argues that the classical AI approach of constraining the search space of possible outcomes by simplifying the world state and action representations may not generate accurate enough projection results to be practically applicable. Instead, he suggests to consider comprehensive world and action representations but restrict the search space to only ``important'' or ``significant'' outcomes.
%
Continuing this line of work, Beetz at al.\ \cite{beetz1997projection} present a plan revision technique that improves the behavior of agents by eliminating probable execution failures.
They estimate the frequency of occurrence of failures and apply plan transformation rules to forestall the most probable ones,
based on running a small number of plan execution samples in projection.
The work demonstrates advanced techniques, however, the application domain of the system is a simulated delivery robot in a 2D grid world,
whereas in the real world domain it is very difficult to construct a realistic probabilistic model of robot's actions and their effects.
In future, we are planning to learn such models using large amounts of robot experience data.

In the robotics community, the idea of using simulators to improve execution is not new.
Rockel et al.\ \cite{rockel2015simulation} show a system where
simulation is integrated into the planner, such that the latter can choose the appropriate action and parameters based on simulation.
This allows the robot to learn a new skill such as balancing an object on a tray.
Kunze et al.\ \cite{kunze2011simulation} present a temporal projection system that translates naive physics problems into parameterized 
simulation tasks with support of first-order representation reasoning over the execution results.
With this system the robot can estimate parameters of actions, e.g., for manipulating an egg.
Abelha et al.\ \cite{tools} use a simulator to estimate how a particular tool performs in a given task:
they wary the parameters of the action of using a tool to estimate the best parameterization 
based on a ``task function''.
%
The difference between the aforementioned works and our approach is that they concentrate on short time span tasks and simulations thereof, whereas our approach implements fast temporal projection over multiple plan steps and can infer a full set of parameters at once.
From a practical perspective, traditional simulation-based approaches are computationally expensive and have a low real-time factor,
whereas our plan projection is very fast with respect to the pace of action execution (see Section \ref{sec:evaluation}).

The closest related work that deals with large time span 
 temporal plan projection is by M{\"o}senlechner et al.\ \cite{mosenlechner2013projection}.
The system described in \cite{mosenlechner2013projection} considers simple sequences of actions designed for
simulation. Our approach aims at complete plans, which run on a real robot and are, therefore, much more complex than those used in simulation.
This requires an approach to structure plans such that they are easily introspected, which is presented in this paper.
Additionally, running projection on a real robot during execution and integrating results of projection-based reasoning back into the executive
poses another challenge, which has been tackled in this paper. 







\section{Plan Architecture}

\begin{figure}[htb]
    \vspace{-1.5ex}
    \centering  
    \includegraphics[width=\columnwidth]{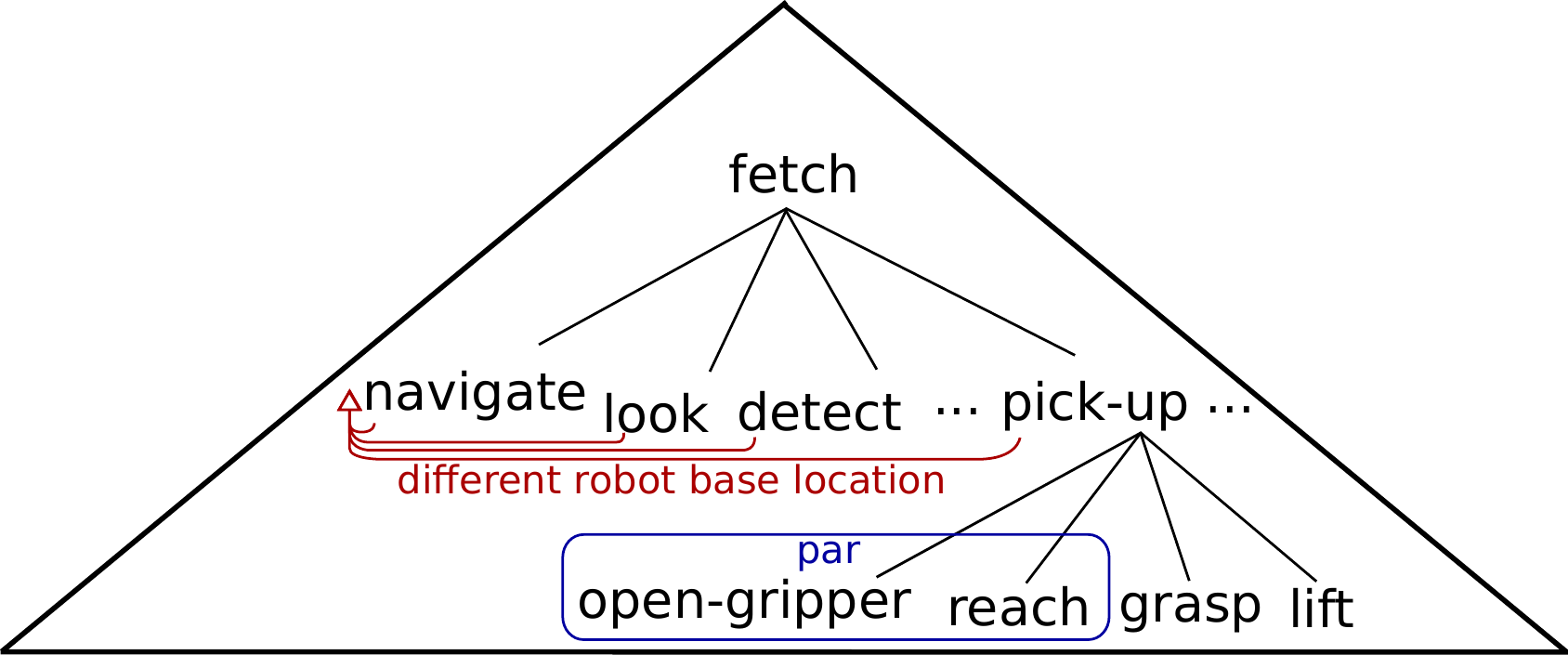}
    \caption{Task tree of a \textit{fetching} action}
    \label{fig:fetch-task-tree}
    \vspace{-1.5ex}
\end{figure}

To enable convenient performance introspection, plans have to be nicely structured, i.e. be modular, explicit and transparent.
We have developed plans for fetching and delivering objects that have such a structure.
In these plans, the control flow is separated from the reasoning necessary to ground entity descriptions into the environment at hand.
The control flow of \textit{fetch} is illustrated in Figure \ref{fig:fetch-task-tree}.


The \textit{fetch} plan consists of sequentially executing four other subplans,
which can generate failures of 5 different types that \textit{fetch} has to be able to handle.
Some of the failures relevant for the fetch and deliver domain are listed in Table \ref{tab:failures}.

\begin{table}[htb]
    \vspace{-0.5ex}
\begin{tabularx}{\columnwidth}{|l|X|}
    \hline
    \textit{perception-object-not-found} & perception system returned no matching object \\
    \hline
    \textit{object-nowhere-to-be-found} & object was not found at the search location despite all failure recovery \\    
    \hline
    \textit{navigation-pose-unreachable} & navigation trajectory is blocked \\
    \hline
    \textit{navigation-pose-in-collision} & navigation goal results in a collision with the environment \\
    \hline
    \textit{navigation-goal-not-reached} & navigation controller finished but goal was not reached \\
    \hline
    \textit{ptu-goal-unreachable} & look goal tries to twist robot's neck \\
    \hline
    \textit{manipulation-pose-unreachable} & no IK solution exists for pose \\
    \hline
    \textit{manipulation-goal-not-reached} & manipulation controller finished but goal was not reached \\
    \hline
    \textit{manipulation-pose-in-collision} & manipulation trajectory generates a collision with the environment \\
    \hline
    \textit{gripper-closed-completely} & gripper closed completely although an object was expected to be grasped \\
    \hline
\end{tabularx}
\caption{Common failures from fetch and deliver domain}
\label{tab:failures}
\vspace{-5ex}
\end{table}

\noindent The default failure recovery strategy of \textit{fetch} is to sample a new
robot base location and retry, as illustrated with the red arrows in Figure \ref{fig:fetch-task-tree}.
The \textit{pick-up} subplan consists of four other subplans, two of which are executed in parallel,
as illustrated with the blue box.
If \textit{fetch} cannot handle a failure locally, it throws an \textit{object-unfetchable} failure to the higher level of the plan hierarchy.

The plans in our system are implemented using the CRAM Plan Language (CPL) \cite{cram},
which is a domain-specific language that provides syntactic sugar for implementing parallelism and synchronization, 
contains failure handling constructs targeted at robotic applications and implements the \textit{entity descriptions} concept mentioned above.
The \textit{fetch} plan, written in CPL, is simple and concise. The knowledge required to
execute the plan successfully in a given environment is inferred through the
reasoning rules for grounding entity descriptions.
Table \ref{tab:knowledge} shows all the knowledge preconditions 
of the \textit{fetch} plan.
These rules define the search space of plan projection, from which the sampling is done.

\begin{table}[htb]
    \vspace{-1.0ex}
\begin{tabularx}{\columnwidth}{|X|}
    \hline
    robot\_base\_location(ReferenceLocations, Robot, Constraints, BaseLoc) \\ \hline
    arm(Object, Robot, Arm) \\ \hline
    grasp\_type(ObjectType, Grasp) \\ \hline
    gripper\_opening(ObjectType, Distance) \\ \hline
    reaching\_trajectory(ObjectType, Arm, GraspType, ObjectPose, Traj) \\ \hline
    grasping\_force(ObjectType, Force) \\ \hline
    lifting\_trajectory(ObjectType, Arm, GraspType, ReachTrajectory, Traj) \\ \hline
\end{tabularx}
\caption{Knowledge preconditions of a fetching action}
\label{tab:knowledge}
\vspace{-5ex}
\end{table}

The reasoning rules are incorporated into the \textit{fetch} plan through its three input parameters: 
the entity descriptions of the object to fetch, the pick up action and the location for the robot to stand when picking up.

\begin{figure}[htb]
    \vspace{-1.0ex}
    \centering  
    \includegraphics[width=\columnwidth]{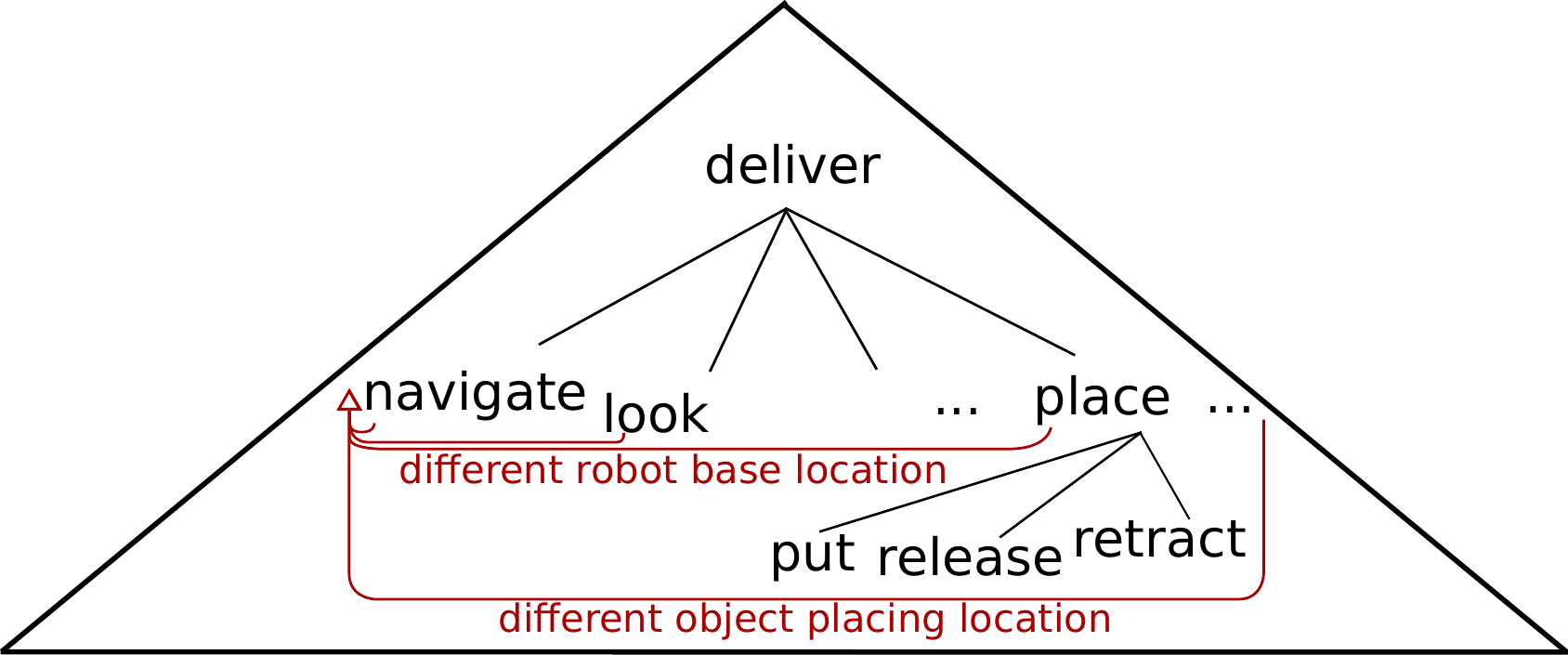}
    \caption{Task tree of \textit{delivering} action}
    \label{fig:deliver-task-tree}
    \vspace{-2ex}
\end{figure}

The \textit{deliver} action is implemented similarly to \textit{fetch}. It is illustrated in Figure \ref{fig:deliver-task-tree}:
it has two hierarchically nested failure recovery strategies for handling four types of failures.

\section{Plan Projection}
\label{sec:projection}

The projection library that we use in our system is the one described in \cite{mosenlechner2013projection}.
It contains functionality for setting up a projection environment where we execute our plans described in previous sections.
To be able to use the library, we provided it with two equivalent implementations of all the low-level motions that the robot can execute
-- one for the real robot and one for projection.
For 3D world representation we use the physics-based geometric world from \cite{bullet}.
%
It uses Bullet physics engine\footnote{\url{http://bulletphysics.org/}} to represent the 3D state of the world and to do physics simulation,
OpenGL's GLUT library\footnote{\url{https://www.opengl.org/resources/libraries/glut/}} to do visibility reasoning and to visualize the Bullet world,
KDL-based\footnote{\url{http://www.orocos.org/wiki/orocos/kdl-wiki}} inverse kinematics solver to do reachability reasoning,
and other external and internal tools.

In projection, all the motions of the robot are not continuous, as 
in traditional simulators, but discrete, so the robot goes through key poses of motions by ``teleporting''. This is the level of abstraction sufficient for our plan projection framework
for making realistic predictions about action outcomes:
physics-based methods provide fine-grained information sufficient to perform geometric reasoning. 
Opposite 
of the precision requirement, 
projection also should not significantly delay execution,
i.e.\ it should be much faster than realtime,
hence, ensuring the correctness of motion controller trajectories is out of its scope.
We assume that low-level controllers generate motions that satisfy the constraints
given by the plan, and in case the controllers throw a failure, those are handled by well-designed failure recovery strategies.
Thus, we achieve modularization and ensure that our plans satisfy the design requirements,
while maintaining necessary accuracy by considering low-level motions through their key poses
(see more related discussion in Section \ref{sec:conclusion}).

We use the same geometric world for robot's belief state representation and for projection.
Due to this tight integration, it is easy to initiate projection with the current belief state of the robot at any point in time
and manipulate it for projecting into the future, then reset it back to the original state representing the real world once projection is over.

We apply the projection library to our carefully designed plans and do performance introspection on them.
As opposed to the typical model-based approach to action planning, where control routines are modeled in a purely symbolic way,
our system represents the control routines in a subsymbolic way but at the same time such that it would be possible
to symbolically infer consequences of executing a plan. This is described in the next section.


\section{Performance Introspection}

\begin{figure}[htb]
    \centering  
    \includegraphics[width=\columnwidth]{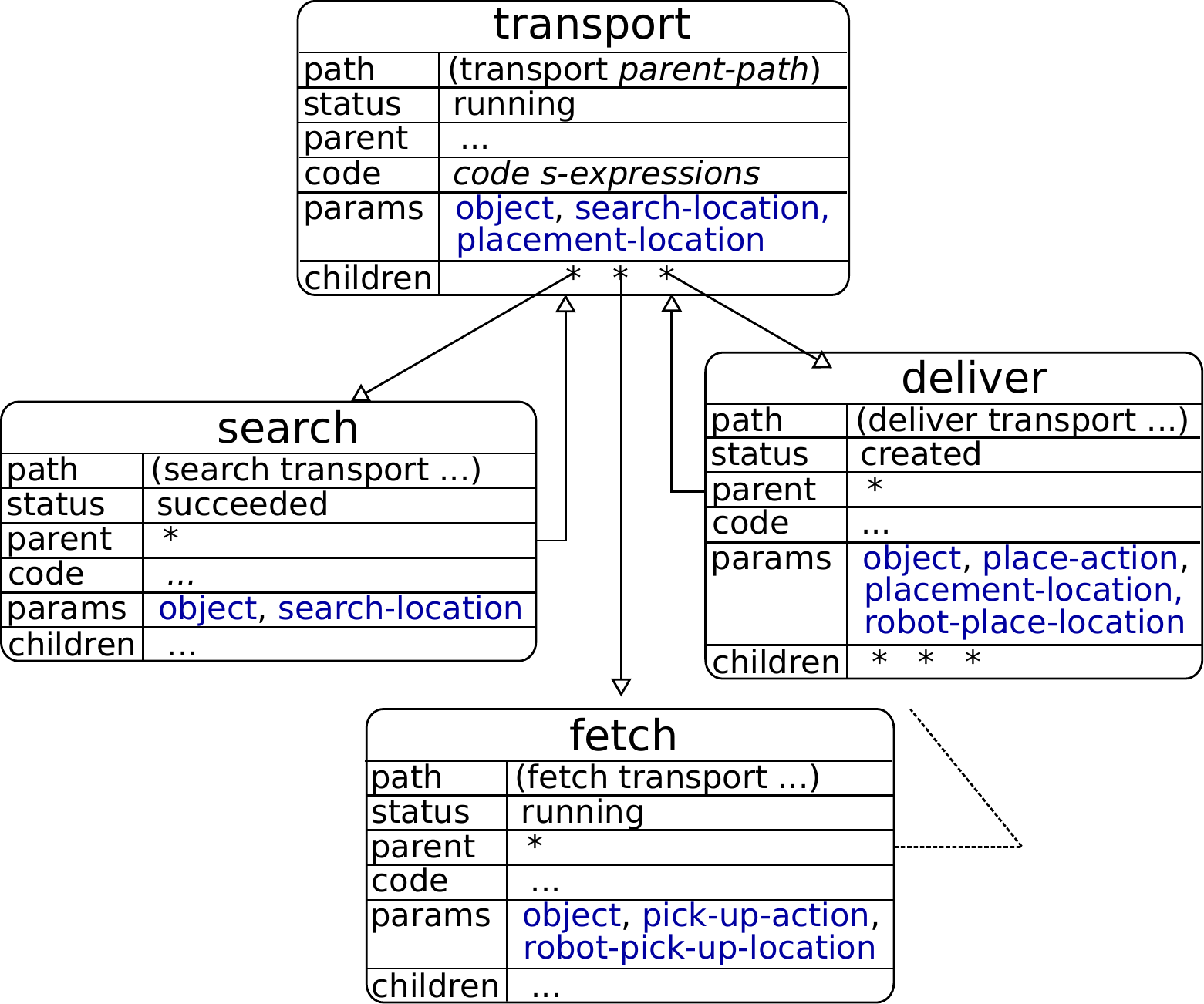}
    \caption{Diagram of the task tree data structure}
    \label{fig:task-tree-datastructure}
\end{figure}

The main data structure in which plan-relevant information is stored during execution is the task tree
(see Figure \ref{fig:task-tree-datastructure}). The nodes of the tree correspond to \textit{tasks}.
A task is a representation of the runtime state of an annotated segment of the robot control program that is semantically meaningful in the context of plan execution and is important for introspection purposes.
The most common task is the representation of an \textit{action description} that is performed within the plan.
Every node in the task tree contains a unique path, which is used for indexing and searching,
a status, which can be any of \textit{created}, \textit{running}, \textit{suspended}, \textit{succeeded}, \textit{evaporated} or \textit{failed},
pointers to the parent node and children nodes, the code expressions of the task, the parameters with which it has been called, information about its failures etc.
%
The task tree is automatically generated at runtime while tasks are being executed.
%
To access the task tree and to reason on it, an API consisting of first-order logic predicates is defined. The ones relevant for this paper are listed in Table \ref{tab:task-tree-rules}.

\begin{table}[htb]
    \vspace{-1.0ex}
\begin{tabularx}{\columnwidth}{|p{0.39\columnwidth}|p{0.54\columnwidth}|}
    \hline
    \textit{task(\_, Task)} & Binds \textit{Task} to any task of the current task tree \\
    \hline
    \textit{task(SubtreePath, Task)} & Binds \textit{Task} to any task of subtree defined with \textit{SubtreePath} \\
    \hline
    \textit{task\_path(Task, Path)} & Gives the unique path of the task node defined with \textit{Task} and binds it to \textit{Path} \\
    \hline
    \textit{task\_outcome(Task, Outcome)} & Binds the result of \textit{Task} to \textit{Outcome} \\
    \hline
    \textit{task\_failure(Task, Failure)} & If \textit{Task} failed, binds its failure object to \textit{Failure} \\
    \hline
    \textit{task\_created\_at(Task, Time)} & Binds the timestamp of creation of \textit{Task} to \textit{Time} \\
    \hline
    \textit{task\_started\_at(Task, Time)} & Binds the timestamp of when \textit{Task} started execution to \textit{Time} \\
    \hline
    \textit{task\_ended\_at(Task, Time)} & Binds the timestamp of when \textit{Task} execution ended to \textit{Time} \\
    \hline
    \textit{action\_subtask(SubtreePath, \_, Task, Action)} & Binds all tasks from \textit{SubtreePath} corresponding to action descriptions to \textit{Task} and their action description to \textit{Action} \\
    \hline
    \textit{action\_subtask(SubtreePath, \nohyphens{ActionType}, Task)} & Binds all tasks from \textit{SubtreePath} corresponding to action descriptions of type \textit{ActionType} to \textit{Task} \\
    \hline   
    \textit{action\_task\_previous\_sibling( \nohyphens{SubtreePath}, Task, ActionType, PrevTask)} & For an action task \textit{Task} in \textit{SubtreePath} finds the previous action task of type \textit{ActionType} and binds it to \textit{PrevTask} \\
    \hline
    \textit{action\_task\_next\_sibling( \nohyphens{SubtreePath}, Task, ActionType, NextTask)} & Binds the next action of type \textit{ActionType} of an action task \textit{Task} in \textit{SubtreePath} to \textit{NextTask} \\
    \hline
\end{tabularx}
\caption{Predicates for accessing task tree data}
\label{tab:task-tree-rules}
\vspace{-5ex}
\end{table}

These predicates can be used by the robot as building blocks for answering
questions such as "What was the last action I was trying to perform?", "Which parameters did I use?", "Was the action successful?", "What were the failures?" etc.
For example, if a placing action failed, the robot could crawl the task tree for the picking up action that preceded the failed placing action to see if the source of failure could have been that the object was picked up in a wrong way.
As the input parameters of tasks are stored in the task tree, the robot can access all the action parameterizations that it used during execution and reason about them by reading out the results of grounding the entity descriptions that were used as parameters of action tasks.
To keep introspection queries simple and straightforward it is crucial for the task tree to be well structured. This is achieved automatically if the plans are designed in a structured and systematic way, as is, for example, the case with our fetching and delivering plans.

\begin{figure}[htb]
    \vspace{-1.5ex}
    \centering  
    \includegraphics[width=0.6\columnwidth]{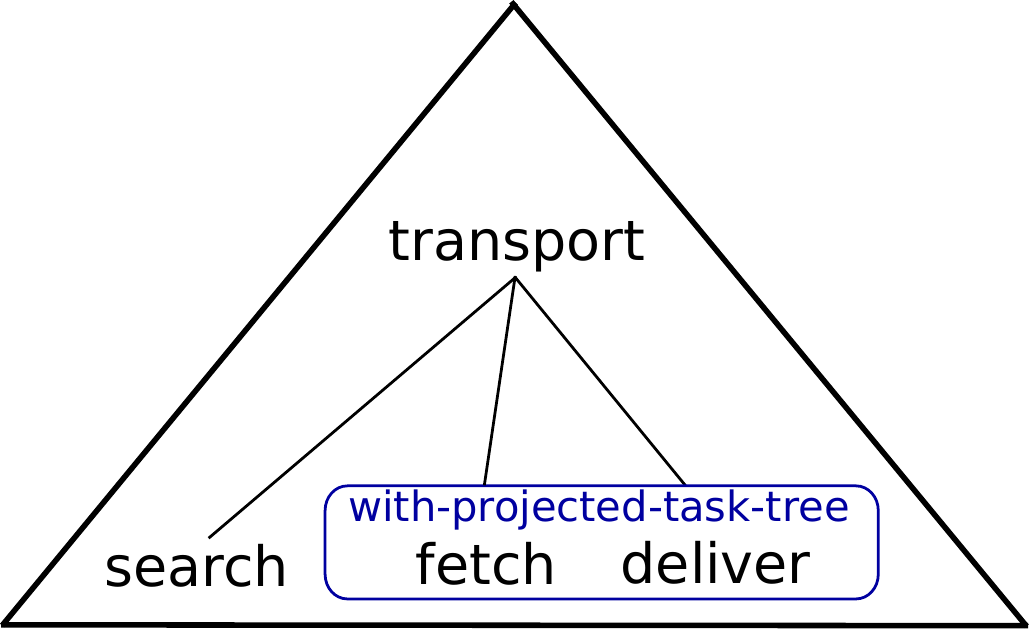}
    \caption{Task tree of \textit{transporting} action}
    \label{fig:transport-task-tree}
    \vspace{-1.5ex}
\end{figure}

Let us consider a transporting action (see Figure \ref{fig:transport-task-tree}). 

We would like to project its plan with different parameterizations of the \textit{fetch} and \textit{deliver} actions and choose the one that leads to successful execution.
We might also want to compare the parameterizations based on a certain cost function, e.g., a function that compares lengths of trajectories the robot would have to execute.
Thus, we use the task tree Prolog API described above in the following way:

\begin{lstlisting}[style=Prolog]
successful_fetch_and_deliver_params(ParentTaskPath,
                  PickNavAction, PickAction,
                  PlaceNavAction, PlaceAction) :-
  action_subtask(ParentTaskPath, fetching,
                 FetchTask, FetchAction),
  task_path(FetchTask, FetchTaskPath),
  action_subtask(FetchTaskPath, picking-up,
                 PickTask, PickAction),
  task_outcome(PickTask, succeeded),
  action_task_previous_sibling(FetchTaskPath,
                               PickTask,
                               navigating,
                               PickNavTask),
  action_subtask(FetchTaskPath, navigating,
                 PickNavTask, PickNavAction),
  action_subtask(ParentTaskPath, delivering,
                 DeliverTask),
  task_outcome(DeliverTask, succeeded),
  task_path(DeliverTask, DeliverTaskPath),
  action_subtask(DeliverTaskPath, placing,
                 PlaceTask, PlaceAction),
  task_outcome(PlaceTask, succeeded),
  action_task_previous_sibling(DeliverTaskPath,
                               PlaceTask, 
                               navigating,
                               PlaceNavTask),
  action_task(DeliverTaskPath, navigating,
              PlaceNavTask, PlaceNavAction).
\end{lstlisting}

We extract the fetching and delivering tasks from the task tree and make sure that their outcomes are \textit{succeeded}.
If not, the rule fails and we do not get any parameter bindings, which means that the projection run was not successful.
Next, we extract the picking up task from the fetching subtree and the action description corresponding to that task.
As we are applying introspection after execution has finished, in case of successful pick up the action description has to
be already grounded. Therefore, we can access all the parameters of that grounding, including the arm that was used, the grasp type,
even the trajectories. Once we have the picking up action, we find the navigating action that last preceded the pick up.
That action contains the location description that was used to position robot's base.

Thus, with a small number of queries we can access all the parameterizations of the general plan that were
used to specialize it to the environment at hand.
This mechanism is only made possible due to the design of plans that we follow,
in which the control flow is separated from the reasoning processes.


\section{Incorporating Projection into Execution}
\label{sec:projection}

To run projection for finding plan parameterizations that lead to successful task execution
we have implemented the \textit{with-projected-task-tree} construct.
It is wrapped around the segment of the robot control program that we would like to project.
For example, in the \textit{transport} plan illustrated in Figure \ref{fig:transport-task-tree}
projection is ran after the \textit{search} action has been executed and the object has been found:
as we would like to optimize the parameters of the picking up action and the placing action,
the location of the object 
has to be known.

The signature of \textit{with-projected-task-tree} is as follows:

\begin{lstlisting}[style=Lisp]
(with-projected-task-tree
   ~entity-descriptions-to-optimize~
   ~number-of-projection-runs~
   ~cost-function-to-compare-results~
 ~code-to-project-and-execute~)
\end{lstlisting}

The transporting plan is, therefore, defined as follows:

\begin{lstlisting}[style=Lisp]
(def-plan transport (~?object~ ~?search-location~
                     ~?delivering-location~)
 (perform (an action
              (type searching)
              (object ~?object~)
              (location ~?search-location~)))

  (@with-projected-task-tree
     (<?fetch-robot-location< &?pick-up-action&
      ^?deliver-robot-location^ !?place-action!)
     4
     #'pick-best-parameters-by-distance@

   (perform 
    (an action
        (type fetching)
        (object ~?object~)
        (robot-location <?fetch-robot-location<)
        (pick-up-action &?pick-up-action&)))

   (perform 
    (an action
        (type delivering)
        (object ~?object~)
        (target ~?delivering-location~)
        (robot-location ^?deliver-robot-location^)
        (place-action !?place-action!)))))
\end{lstlisting}

The code segment with fetching and delivering actions will be executed in projection four times and the resulting four parameters will be compared with the \textit{pick-best-parameters-by-distance} cost function.
Finally, the same code segment will be executed on the real robot with parameters from the best projection run.


\section{Experimental Analysis}
\label{sec:evaluation}

\begin{figure}[htb]
    \centering  
    \includegraphics[width=\columnwidth]{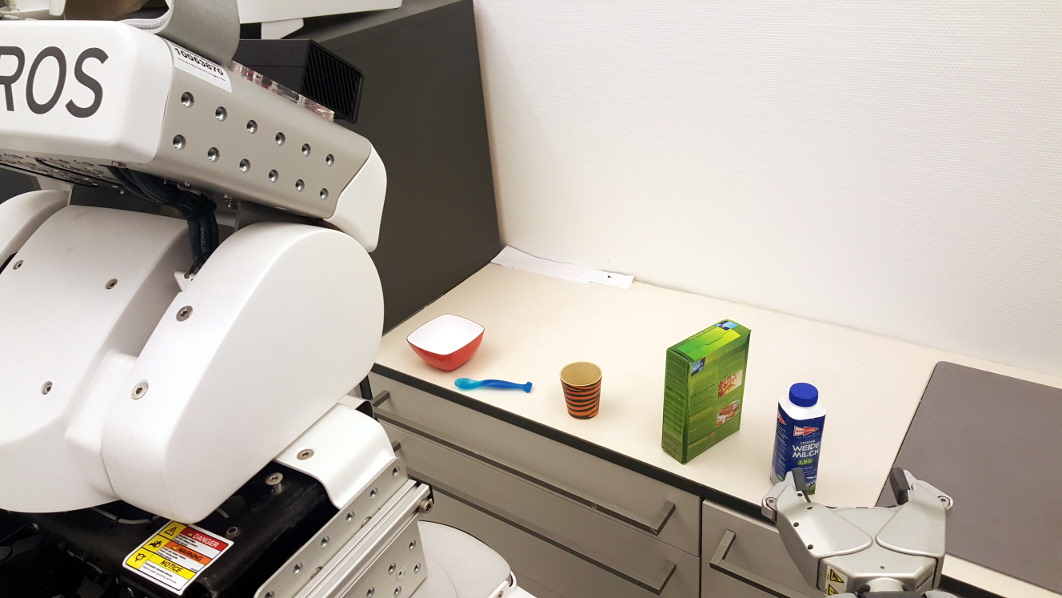}
    \caption{Experimental setup -- initial configuration}
    \label{fig:experimental-setup}
    \vspace{-3ex}
\end{figure}

We evaluated our approach on a breakfast table setting scenario with a PR2 robot.
The scenario included fetching 5 different objects and bringing them to the table.
We executed it 10 times without our system and 10 times with it.
In an effort to reduce the randomness factor in execution we constrained
the initial as well as goal locations of objects to be constant in all the runs.
The initial configuration we chose was random, with the constraint that
objects should be at least 2 cm away from each other and not be completely out
of reach of the robot. The setup is shown on Figure \ref{fig:experimental-setup}.
The robot transports the objects one by one in the following order:
\textit{milk}, \textit{cup}, \textit{cereal}, \textit{bowl}, \textit{spoon}.

The first action in the \textit{transport} plan is the searching action,
so the robot searches for the object of a specific type on the surface of the counter.
As it does not know where exactly the object is and as its field of view
is limited to the sensor's image size, it samples random poses on the surface,
navigates to a location from where the pose is visible, and moves its head to point
at it. Then it calls the perception system \cite{robosherlock}. If perception fails, the robot
picks a different pose on the surface and retries.

Once the object has been found, next in the plan are the \textit{fetch} and \textit{deliver} actions wrapped into
\textit{with-projected-task-tree} as shown in Figure \ref{fig:transport-task-tree}.
If projection is disabled, the robot samples a location to stand to reach the object, drives to the location,
samples an arm and a grasp type to use and tries to reach. If there is no
IK solution or there is an occlusion, a manipulation failure occurs, so the robot samples a different location to stand, drives there and retries.
This backtracking behavior is time consuming and leaves an impression of incompetent behavior.
Additionally, if the object placing orientation is difficult to achieve with a certain grasp and
the robot is unlucky to sample that particular grasp, \textit{deliver} will completely fail.
If projection is enabled, the robot executes four runs of projection.
We refer to \cite{beetz1997projection} for the proof of an argument that with a small number of randomly generated execution scenarios it is nonetheless very likely that the probable failures will be eliminated.
Projection is used to choose the following four parameters: the arm to grasp with, the grasp type and
the locations of the robot base for picking up and placing the objects.
Action parameterizations in successful projection runs are evaluated based on a heuristic approximation of distances that the robot
would have to move and, thus, the best run is chosen.
Chosen plan parameters are then used in the real world to execute the \textit{fetch} and \textit{deliver} actions.

Our perception system has about 3 cm precision for the objects
in the experimental setup, which tends to improve when the robot gets closer to the object.
Due to that, the fetching plan reperceives the object 
directly before grasping. We were faced with two alternatives:
either perceive the object once, project, then reuse generated trajectories even if the result of reperception varies highly from the previous result,
or not reuse the trajectories but rather only predict the arm and the grasp type that is most likely to succeed.
We went in favor of the second, as reusing trajectories proved to be very prone to misgrasping errors.

As the aim of projection is to find the action parameterization that leads to successful
task execution, we chose our evaluation criteria to be the success rate of actions
and the number of failures that happen.
The types of most prominent failures that decided the outcome of actions
were the ones related to reachability, including \textit{object-unreachable},
\textit{manipulation-pose-unreachable} and \textit{navigation-pose-unreachable},
and the environment collision failures, including \textit{manipulation-pose-in-collision}
and \textit{navigation-pose-in-collision}.
Tables below show statistics from the first run of the system without using projection (Table \ref{tab:np-experiment-one})
and the first run with projection (Table \ref{tab:wp-experiment-one}).

\begin{table}[htb]
    \vspace{-1.5ex}
\begin{tabularx}{\columnwidth}{|l||X|X|X|X|X|X|}
    \hline
    \textbf{Object}       & \textbf{milk} & \textbf{cup} & \textbf{cereal} & \textbf{bowl} & \textbf{spoon} & \textbf{Total} \\ \hline
    \textbf{Runtime}      & 411.6         & 207.2        & 142.4           & 170.9         & 229.2          & 1181.4         \\ \hline
    \textbf{Arm used}     & left          & right        & right           & right         & right          &                \\ \hline
    \textbf{Grasp used}   & front         & front        & back            & top           & top            &                \\ \hline
    \textbf{Success}      & no            & yes          & yes             & yes           & yes            & 4 of 5         \\ \hline
    \textbf{Coll. fail.}  & 7             & 9            & 4               & 3             & 4              & 27             \\ \hline
    \textbf{Reach. fail.} & 45            & 13           & 10              & 0             & 0              & 68             \\ \hline
    \textbf{Sum failures} & 52            & 22           & 14              & 3             & 4              & 95             \\ \hline
\end{tabularx}
\caption{Experimental results of one scenario run without projection}
\label{tab:np-experiment-one}
\vspace{-4ex}
\end{table}

In Table \ref{tab:np-experiment-one} it can be seen that the delivering action of the milk object failed
and created 52 collision and reachability failures, which resulted in applying failure handling strategies that relocate robot's base.
%
The average number of failures that happened in the 10 scenario runs that did not use projection, as seen in Table \ref{tab:np-experiments-avg}, is 55.45 failures per run.

\begin{table}[htb]
    \vspace{-1.5ex}
\begin{tabularx}{\columnwidth}{|l||X|X|X|X|X|X|X|}
    \hline
    \textbf{Object}       & \textbf{milk} & \textbf{cup} & \textbf{cereal} & \textbf{bowl} & \textbf{spoon} & \textbf{Total} & \textbf{Per obj.} \\ \hline
    \textbf{Coll. fail.}  & 5             & 6.2          & 1.8             & 2.6           & 6.8            & 22.4          &  4.48  \\ \hline
    \textbf{Reach. fail.} & 17.25         & 2.8          & 10              & 2.8           & 0.2           &  33.05          & 6.61 \\ \hline
    \textbf{Total fail.}  & 22.25         & 9.0          & 11.8            & 5.4           & 7.0           &  55.45          &  11.09\\ \hline
    \textbf{Success rate} & 75\%          & 100\%        & 100\%           & 100\%         & 80\%          &   91\%          &  91\%  \\ \hline
\end{tabularx}
\caption{Results averaged over ten scenario runs without projection}
\label{tab:np-experiments-avg}
\vspace{-4ex}
\end{table}

Experimental results of one scenario run with the projection system enabled are shown in Table \ref{tab:wp-experiment-one}.
It can be seen that the number of reachability and collision failures is very small.
There is one collision failure that happened when transporting the \textit{spoon} object.
However, it was expected to be 0 if the parameterization was predicted to generate successful behavior.
As mentioned before, we only infer the arm to use and the grasp type, which should lead to successful execution,
and do not reuse the trajectory generated in projection.
As the trajectory generation algorithm is randomized, even slight changes in
object pose can result in no valid trajectory being found.
The collision failure happened when picking up the spoon because the perception system
changed the pose estimate of the object significantly enough for the inverse kinematics solver to fail for the new pose
with the given arm and grasp type. This poses an issue for the projection mechanism if the perception results
are inconsistent with respect to object orientations, which is the case in our perception system:
the axis of the object pose can flip randomly. In that case, e.g., a front grasp that was
supposed to be ideal for the current world state becomes unreachable and a back grasp should be chosen instead.
This situation happened in one of the 10 runs of the scenario with projection, where
the grasp for the milk object failed although a valid parameterization was successfully inferred.

\begin{table}[htb]
    \vspace{-1.5ex}
\begin{tabularx}{\columnwidth}{|l||X|X|X|X|X|X|}
    \hline
    \textbf{Object}       & \textbf{milk} & \textbf{cup} & \textbf{cereal} & \textbf{bowl} & \textbf{spoon} & \textbf{Total} \\ \hline
    \textbf{Proj. time}   & 47.9          & 25.0         & 23.3            & 12.4          & 15.5           &                \\ \hline
    \textbf{Infer. time}  & 31.9          & 5.6          & 4.2             & 3.3           & 3.8            &                \\ \hline
    \textbf{Runtime}      & 193.8         & 155.6        & 151.5           & 132.5         & 160.7          & 823.2          \\ \hline
    \textbf{Arm used}     & right         & right        & right           & right         & right          &                \\ \hline
    \textbf{Grasp used}   & front         & front        & back            & top           & top            &                \\ \hline
    \textbf{Proj. success}& 2 of 4        & 4 of 4       & 4 of 4          & 4 of 4        & 4 of 4         &                \\ \hline
    \textbf{Success}      & yes           & yes          & yes             & yes           & yes            & 5 of 5         \\ \hline
    \textbf{Coll. fail.}  & 0             & 0            & 0               & 0             & 1              & 1              \\ \hline
    \textbf{Reach. fail.} & 0             & 0            & 0               & 0             & 0              & 0              \\ \hline
    \textbf{Sum failures} & 0             & 0            & 0               & 0             & 1              & 1              \\ \hline
\end{tabularx}
\caption{Experimental results of one scenario run with projection}
\label{tab:wp-experiment-one}
\vspace{-4ex}
\end{table}

Table \ref{tab:wp-experiments-avg} shows the average number of failures that happen in real world when using fast plan projection.

\begin{table}[htb]
    \vspace{-1.5ex}
\begin{tabularx}{\columnwidth}{|l||X|X|X|X|X|X|X|}
    \hline
    \textbf{Object}       & \textbf{milk} & \textbf{cup} & \textbf{cereal} & \textbf{bowl} & \textbf{spoon} & \textbf{Total} & \textbf{Per obj.} \\ \hline
    \textbf{Coll. fail.}  & 2.1           & 1.0          & 0.3             & 0.0           & 0.2            & 3.6           &  0.72  \\ \hline
    \textbf{Reach. fail.} & 4.8           & 1.22         & 0.2             & 0.0           & 0.0           &  6.22          & 1.24\\ \hline
    \textbf{Total fail.}  & 6.9           & 2.22         & 0.5             & 0.0           & 0.2           &  9.82         &  1.96\\ \hline
    \textbf{Success rate} & 80\%          & 100\%        & 100\%           & 100\%         & 100\%          &   96\%          &  96\%  \\ \hline
\end{tabularx}
\caption{Results averaged over ten scenario runs with projection}
\label{tab:wp-experiments-avg}
\vspace{-4ex}
\end{table}

Based on experimental data we can conclude that our system improves the success rate of fetch and deliver plans from 90\% to 96\%, which is not substantial since the robustness of the evaluation scenario is already considerably high. However, we additionally decrease the amount of manipulation-related failures and, therefore, times when robot physically backtracks, from 55.45 per run to 9.82, which is more than a 500\% improvement (see Figure 7).

\begin{figure}[htb]
    \vspace{-1.5ex}
\begin{tikzpicture}
    \begin{groupplot}[group style={
                      group name=myplot,
                      group size= 2 by 1,
                      horizontal sep=1.8cm}]
        \nextgroupplot[
            width=0.5\columnwidth,
            height=0.5\columnwidth,
            ybar,
            legend style={at={(0.5,-0.25)},
              anchor=north,legend columns=-1},
            ylabel={success rate (\%)},
            y label style={at={(-0.25,0.5)}},
            ymin=0, ymax=120,
            symbolic x coords={avg. over 10 runs},
            xtick=data,
            nodes near coords,
            nodes near coords align={vertical},
            ]
        \addplot[draw=red,fill=red!10] coordinates {(avg. over 10 runs,91)};\label{plots:plot1}
        \addplot[draw=blue,fill=blue!60] coordinates {(avg. over 10 runs,96)};\label{plots:plot2}
                
        \nextgroupplot[
            width=0.5\columnwidth,
            height=0.5\columnwidth,
            ybar,
            legend style={at={(0.5,-0.25)},
              anchor=north,legend columns=-1},
            ylabel={manip. failures},
            y label style={at={(-0.25,0.5)}},
            ymin=0, ymax=100,
            symbolic x coords={avg. over 10 runs},
            xtick=data,
            nodes near coords,
            nodes near coords align={vertical},
            ]
        \addplot[draw=red,fill=red!10] coordinates {(avg. over 10 runs,55.45)};
        \addplot[draw=blue,fill=blue!60] coordinates {(avg. over 10 runs,9.82)};
    \end{groupplot};
\path (myplot c1r1.south west|-current bounding box.south)--
      coordinate(legendpos)
      (myplot c2r1.south east|-current bounding box.south);
\matrix[
    matrix of nodes,
    anchor=north,
    draw,
    inner sep=0.2em,
    draw
  ]at([yshift=-0ex]legendpos)
  {
    \ref{plots:plot1}& without projection &[5pt]
    \ref{plots:plot2}& with projection &\\};
\end{tikzpicture}
    \vspace{-1.0ex}
\label{fig:histogram}
\caption{Success rate and number of manipulation failures comparison between executions without and with projection.}
    \vspace{-1ex}
\end{figure}
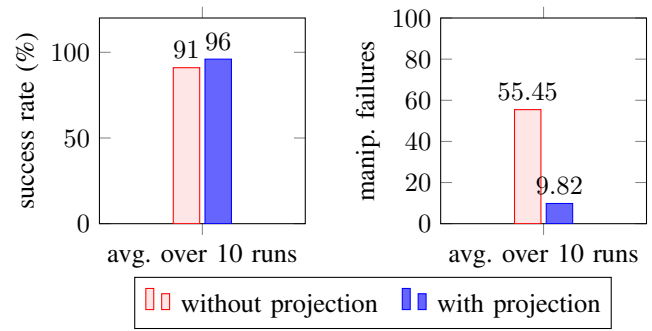

Due to the high variety, we do not consider execution time an important evaluation criteria.
In the 10 times we ran the scenario with projection-based reasoning disabled, execution time varied between 696.4s and 1181.4s.
There is a vast number of factors that affect the runtime of execution on a robotic system,
including the efficiency of computational processes, power of the underlying hardware, time optimality of robot's controllers etc.,
as well as the amount of physical backtracking that happens due to, e.g., perception failures or suboptimal action parameterization samples.
To give a meaningful empirical estimate of execution time, the number of runs would have to be significantly higher than what we have. Considering that
one scenario run takes about 15min, a large-scale evaluation might not be feasible.
However, we do mention some statistics below to give a rough idea of our implementation efficiency for practical reasons.
In the 10 runs of the scenario with projection enabled, 196 projection runs have been executed.
The average projection time per one run of transporting one object was 6.2s,
and 4.1s if excluding all the non-successful runs.
This can be considered sufficiently fast with respect to the pace of action execution.
The hardware used for projection was a laptop with 8GB RAM, an 8 core i7 CPU and an NVIDIA GeForce GT 650M graphics card.
The average runtime of a full scenario run was 877.28s without projection and 823.85s with projection.



\section{Conclusion, Discussion and Future Work}
\label{sec:conclusion}

In this paper we presented the fast plan projection mechanism, which can be used to specialize a general plan towards the environment and task at hand by
choosing action parameterizations that are predicted to lead to successful task execution.
We showed how carefully designed plan structure can benefit plan introspection and how to apply introspection tools to choose parameterizations of executed actions
that were predicted to succeed in the projection environment. We demonstrated how the results are easily integrated into robot's executive such that the optimized part of the plan can be executed right away in the real world. Finally, we evaluated our approach by showing how a PR2 robot is able to use the system to choose action parameterizations that increase task execution success rates and decrease failure rates of fetch and deliver actions in a real world setting.

In the evaluation section we mentioned one limitation of our approach, which is a general limitation that any system that thinks ahead of time based on the current belief state has: if the belief state representation is inaccurate, projection has a higher chance of producing action parameterizations, which do not lead to successful task execution when transferred onto the real world. We overcome this problem by integrating our projection results only as a suggestion for the planner, and if suggested parameterization fails, execution continues  with its default failure handling routines, trying to find a better parameterization without the help of projection-based reasoning. A similar limitation is the danger of the belief state changing while projection performs its inference. In our application scenarios, which happen in semi-controlled environments, external influence and, therefore, unexpected belief state changes happen with a sufficiently low frequency compared to the runtime of the projection-based inference.

One important assumption that has to be made about the projection mechanisms is that the probability distribution of failing in projection is similar to the real world. In our fetch and deliver scenarios it is the case as most failures happen due to (1) inverse kinematics solver not finding a solution, which is the exact same mechanism in projection and in real world, and (2) the goal poses being unreachable due to collisions with the environment, which are also realistic in the projection environment as we use a high-precision model of the environment. 
However, some other failures such as an object slipping away from the gripper, are not represented in our system, and that is a limitation. In future, we are planning to learn failure models to use in our projection environment based on data collected from real world experiments.

\section*{Acknowledgments}
\begin{small}
\noindent
This work was supported by DFG Collaborative Research Center \emph{Everyday Activity Science and Engineering (EASE)} (CRC \#1320).
\end{small}

\bibliography{literature}
\bibliographystyle{IEEEtran}
\end{document}